# Physics-Informed Neural Network Modeling of Biodegradable Contaminant Transport through GCL/SL Composite Liners

Dong Li[1], Yapeng Cao[2*], Haiping Zhao[1], Shutong Han[3]

## ABSTRACT

This study develops a two-domain physics-informed neural network framework for contaminant transport through a GCL/SL composite liner system, in which the thin GCL layer is treated using a steady-state advection-dispersion-biodegradation formulation and the underlying soil liner is modeled as a transient transport domain. Two formulations are evaluated against analytical and finite-element reference solutions under different leachate-head conditions: a standard PINN with soft constraint enforcement (Std-PINN) and a hard-constrained PINN (H-PINN), in which selected boundary and initial conditions are embedded directly into the trial solutions. The Std-PINN captures the overall breakthrough behavior but shows larger errors during the early transport stage, particularly under higher leachate heads where advective transport becomes more pronounced. The H-PINN reduces the optimization burden associated with penalty-based constraint enforcement and provides more accurate and stable concentration predictions, lowering the MAE from approximately 0.058-0.067 for the Std-PINN to about 0.011-0.023 for the H-PINN, while reducing the MRE from approximately 9.10%-19.16% to about 2.08%-3.14%. Parametric analyses confirm that the H-PINN with the tanh activation function and an optimized network structure provides the best predictive accuracy. The H-PINN is further extended to inverse modeling for identifying the SL degradation half-life from limited concentration observations, showing reliable convergence toward prescribed values and acceptable robustness under low-to-moderate observation noise.

---

[1] PhD, Department of Civil, Environmental, and Infrastructure Engineering, George Mason University, Fairfax, VA 22030, USA

[2*] PhD, State Key Laboratory of Cryospheric Science and Frozen Soil Engineering, Northwest Institute of Eco-Environment and Resources, Chinese Academy of Sciences, Lanzhou 730000, China; Laboratoire Navier/CERMES, École Nationale des Ponts et Chaussées, Institut Polytechnique de Paris, 77455 Marne-la-Vallée cedex 2, France (Corresponding author) email: caoyapeng@lzb.ac.cn

[3]MS, Department of Civil and Environmental Engineering, the University of New South Wales, Sydney, Australia

# 1. Introduction

Composite liners are generally installed at the bottom of landfills to protect the surrounding environment and groundwater from potential contamination caused by landfill leachate (Kalbe et al., 2002; Du et al., 2009; Rowe, 2011; Varank et al., 2011; Rowe et al., 2023). A conventional composite liner system commonly consists of a geomembrane (GMB) overlying a low-permeability compacted clay liner (CCL) (Rowe and Booker, 2005; Dickinson and Brachman, 2008; Du et al. 2009; Zhao et al., 2025). In modern landfill applications, geosynthetic clay liners (GCLs) are frequently used as an alternative or supplement to CCLs because they can provide comparable or lower hydraulic conductivity while offering easier installation, improved construction quality control, and reduced consumption of landfill airspace due to their relatively thin thickness (typically 5-10 mm) (Jo et al., 2001, 2005; Kolstad et al., 2004; Bradshaw et al., 2013; Li et al., 2024, 2025). However, because GCLs are relatively thin, their contaminant storage, adsorption, and degradation capacities are limited. Therefore, GCLs are commonly combined with an underlying soil liner (SL), which provides additional thickness, sorption capacity and attenuation potential, thereby delaying contaminant breakthrough and improving the long-term containment performance of landfill liner systems (Foose, 2010; Guyonnet et al., 2001; Koerner, 2008). Contaminants may migrate through composite liners by leakage through geomembrane defects and by diffusion through intact geomembranes, with defect leakage being strongly affected by leachate head and defect density, while diffusive transport through intact geomembranes is governed mainly by concentration gradients and geomembrane-contaminant interactions (Rowe and Booker, 2005; Dickinson and Brachman, 2008; Du et al., 2009; Foose, 2010). Once contaminants enter the underlying GCL/SL barrier system, their migration is further controlled not only by advection, diffusion, and sorption, but also by attenuation processes such as biodegradation. Organic contaminants have been shown to undergo significant degradation in landfill liners, and biodegradation and chemical transformation within liner materials may further influence the breakthrough behavior (Sawhney and Kozloski, 1984; Angelidaki et al., 2000; Weber et al., 2011; Olasupo et al., 2024).

Both numerical and analytical methods have been developed to model contaminant advection and diffusion in composite liner systems (Liu and Ball, 1998; Foose et al., 2002; Chen et al., 2009; Li and Cleall, 2011; Guan et al., 2014; Wu et al., 2016; Feng et al., 2019). Liu and Ball (1998) presented an analytical solution for the solute diffusion in semi-infinite two-layer porous media using a Laplace-transformed Green's function approach, and the solution showed good agreement with finite-difference results. Chen et al. (2009) and Wu et al. (2016) developed analytical solutions for one-dimensional contaminant diffusion in multilayered systems under steady-state flow conditions, with Wu et al. (2016) further incorporating liner degradation effects. Li and Cleall (2011) presented analytical solutions for advective-dispersive solute transport in double-layered finite porous media under different combinations of fixed concentration, fixed flux, and zero-concentration-gradient boundary conditions. Guan et al. (2014) developed an analytical solution for solute advection-dispersion transport through a GCL/SL system considering degradation in both the GCL and SL, although solute transport through the GCL was assumed to be steady state. Feng et al. (2019) further developed a fully transient analytical solution for degradable organic contaminant transport through a composite liner consisting of a geomembrane (GMB), geosynthetic clay liner (GCL), and soil liner (SL).

Although analytical solutions can provide exact or semi-analytical benchmarks, they are typically restricted to idealized geometries, simplified boundary conditions, homogeneous or piecewise-homogeneous material properties, and linear transport processes. These assumptions may limit their applicability to practical composite liner systems, where contaminant transport can be affected by layer-dependent properties, transient boundary conditions, degradation, sorption, and interface continuity between different liner components. Numerical methods, such as the finite-difference method and finite-element method, provide a more flexible approach for approximating the governing partial differential equations under complex initial, boundary, and material conditions (Rowe and Booker, 2005; Zhang et al., 2013; Feng et al., 2019). However, conventional numerical methods generally require mesh generation, discretization, and repeated simulations for parametric or inverse analyses, which may increase computational effort when a large number of scenarios must be evaluated. In addition, inverse identification of transport or degradation parameters from sparse concentration observations often require repeated forward simulations and may be sensitive to discretization choices and measurement noise.

Recently, physics-informed neural networks (PINNs) and related constrained neural-network methods have gained increasing attention as mesh-free approaches for solving partial differential equations (PDEs) (Raissi et al., 2019; He and Tartakovsky, 2021; Li et al., 2026). The original PINN framework introduced by Raissi et al. (2019) incorporates governing-equation residuals and initial/boundary conditions into the neural-network loss function, allowing neural networks to solve forward and inverse PDE problems while respecting known physical laws. He and Tartakovsky (2021) applied PINNs to forward and backward advection-dispersion equations coupled with Darcy flow, demonstrating their potential for contaminant transport modeling. Saadat et al. (2022) investigated the training behavior of PINNs for linear advection-diffusion equations using neural tangent kernel theory and showed that PINNs may experience training difficulties due to spectral bias and imbalance among different loss components. Faroughi et al. (2023) investigated PINNs with periodic activation functions for transient solute transport in heterogeneous porous media, while Berardi et al. (2025) extended inverse PINNs to diffusion, advection-diffusion-reaction, and mobile-immobile transport models in porous materials. Ke et al. (2025) developed pre-trained PINNs for forward and inverse contaminant transport analysis in soils, incorporating pre-training, uncertainty quantification, and domain-decomposition strategies Despite these advances, existing PINN-based transport studies have mainly focused on homogeneous or simplified porous-media systems. Limited attention has been given to contaminant transport through layered GCL/soil liner composite systems, where layer-dependent advection, dispersion, retardation, biodegradation, and material-interface continuity must be considered simultaneously. In particular, the concentration and mass-flux continuity conditions at the GCL/SL interface introduce additional physical constraints that are rarely addressed in existing PINN transport models. This issue is especially important for GCL/SL composite liners because the sharp contrast in hydraulic, diffusive, and degradation properties between the GCL and SL can produce strong concentration gradients and discontinuities in transport coefficients near the material interface.

Therefore, this study develops a two-domain physics-informed neural network (PINN) framework for biodegradable contaminant transport through a GCL/SL composite liner system. Both a standard soft-constrained PINN (Std-PINN) and a hard-constrained PINN (H-PINN) are formulated and evaluated. The proposed models are validated against analytical and finite-element reference solutions under different leachate-head conditions. In addition, parametric analyses are conducted to investigate the effects of activation function, neural-network structure, and SL biodegradation half-life on model performance and breakthrough behavior. Finally, the H-PINN is extended to inverse modeling for identifying the SL degradation half-life from limited concentration observations and assessing its robustness under noisy data.

## 2. Governing Equations

As shown in Fig. 1, the contaminant is assumed to migrate downward through a saturated two-layer composite liner system consisting of a geosynthetic clay liner (GCL) and an underlying soil liner (SL). The vertical coordinate $z$ is measured positive downward from the top surface of the GCL. Thus, $z = 0$ corresponds to the GCL/leachate interface, $z = L_{\mathrm{GCL}}$ corresponds to the GCL/SL interface, and $z = L_{\mathrm{GCL}} + L_{\mathrm{SL}}$ corresponds to the bottom of the soil liner. $L_{\mathrm{GCL}}$ and $L_{\mathrm{SL}}$ are the thicknesses of the GCL and SL, respectively. The hydraulic leachate head above the GCL is denoted as $H_w$, and the contaminant concentration in the overlying leachate is assumed to be constant at $C_0$.

The computational domain can be expressed as:

$$\Omega_{\mathrm{GCL}} = \{z | 0 < z \leq L_{\mathrm{GCL}}\} \tag{1}$$

$$\Omega_{\mathrm{SL}} = \{z | L_{\mathrm{GCL}} < z < L_{\mathrm{GCL}} + L_{\mathrm{SL}}\} \tag{2}$$

where $\Omega_{\mathrm{GCL}}$and $\Omega_{\mathrm{SL}}$denote the GCL and SL domains, respectively.

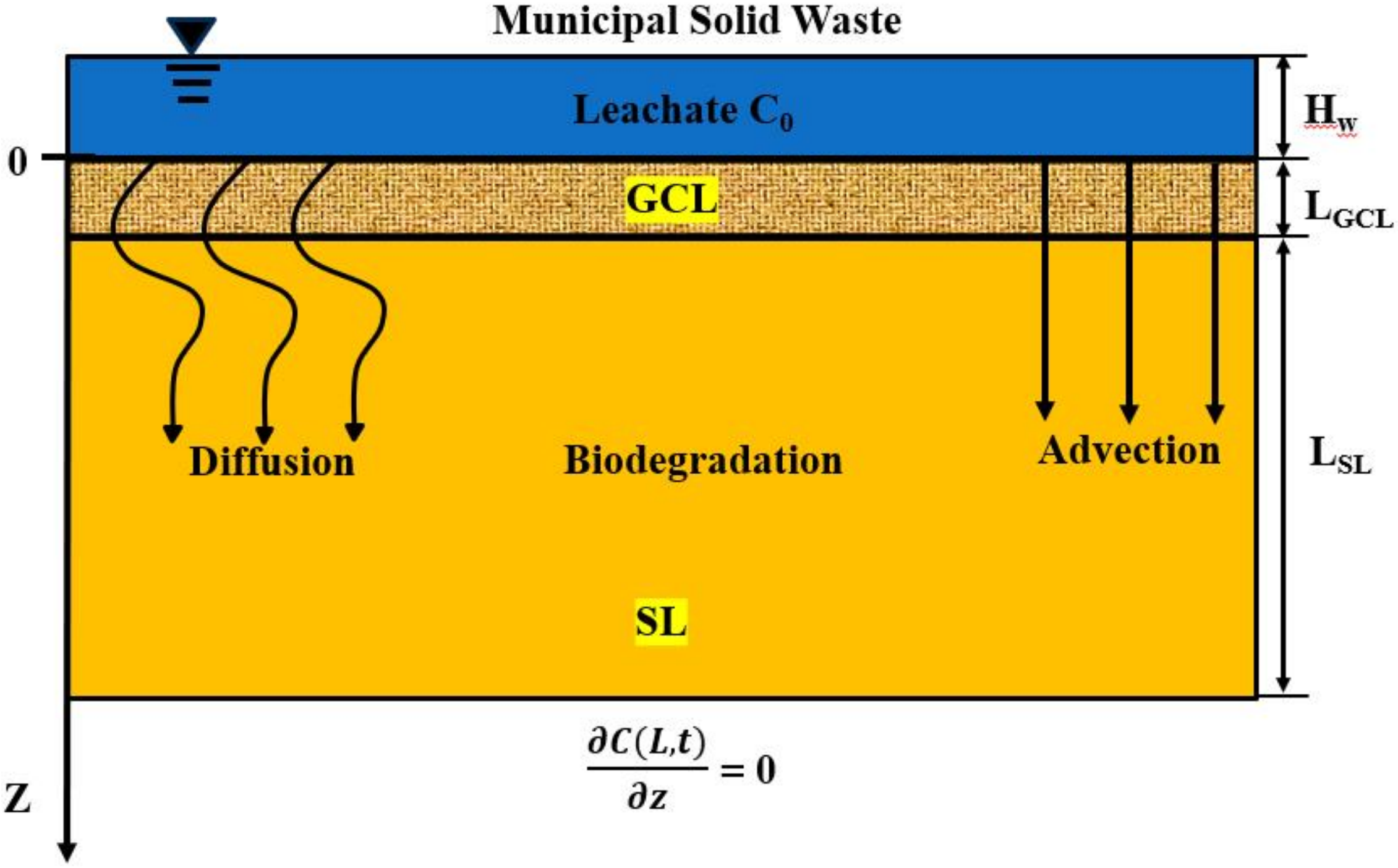

**Fig. 1.** Schematic diagram of transient contaminant transport through a GCL/SL composite system considering advection, diffusion, and biodegradation.

Because the GCL is much thinner than the underlying soil layer, solute transport within the GCL can be approximated as a steady-state process. Therefore, no initial condition is imposed within the GCL layer. For the GCL layer ($0 < z \leq L_{\mathrm{GCL}}, t > 0$), the steady advection-dispersion-biodegradation equation is written as:

$$D_{\mathrm{GCL}} \frac{\partial^2 C_{\mathrm{GCL}}}{\partial z^2} - v_{\mathrm{GCL}} \frac{\partial C_{\mathrm{GCL}}}{\partial z} - \lambda_{\mathrm{GCL}} R_{d,\mathrm{GCL}} C_{\mathrm{GCL}} = 0 \tag{3}$$

where $C_{\mathrm{GCL}}(z,t)$ represents the pore-water solute concentration within the GCL at any position z and any time t, $D_{\mathrm{GCL}}$ refers to the hydrodynamic dispersion of GCL, $v_{\mathrm{GCL}}$ is seepage velocity through the GCL, and $\lambda_{\mathrm{GCL}}$ denotes the decay constant of the solute in GCL. Similarly, for the SL layer ($L_{\mathrm{GCL}} < z < L_{\mathrm{GCL}} + L_{\mathrm{SL}}, t > 0$), the governing equation is:

$$R_{d,\mathrm{SL}} \frac{\partial C_{\mathrm{SL}}}{\partial t} = D_{\mathrm{SL}} \frac{\partial^2 C_{\mathrm{SL}}}{\partial z^2} - v_{\mathrm{SL}} \frac{\partial C_{\mathrm{SL}}}{\partial z} - \lambda_{\mathrm{SL}} R_{d,\mathrm{SL}} C_{\mathrm{SL}} \tag{4}$$

where $C_{\mathrm{SL}}(z,t)$, $D_{\mathrm{SL}}$, $v_{\mathrm{SL}}$, $R_{d,\mathrm{SL}}$, and $\lambda_{\mathrm{SL}}$ are the pore-water solute concentration, hydrodynamic dispersion, seepage velocity, retardation factor, and decay constant of the solute in the soil liner.

The retardation factor for each layer is defined as:

$$R_{d,i} = 1 + \frac{\rho_{d,i} K_{d,i}}{n_i}, i = \mathrm{GCL}, \mathrm{SL} \tag{5}$$

where $\rho_{d,i}$ is the dry density, $K_{d,i}$ is the distribution coefficient, and $n_i$ is the porosity of GCL or SL liner.

The decay constant $\lambda_{\mathrm{GCL}}$, $\lambda_{\mathrm{SL}}$ can be expressed as a function of the half-life $t_{1/2}$ for a first-order biodegradation process:

$$\lambda_i = \frac{\ln 2}{t_{1/2,i}}, i = \mathrm{GCL}, \mathrm{SL} \tag{6}$$

where $t_{1/2,i}$ is the contaminant half-life in GCL or SL liner.

The composite liner is assumed to contain no contaminant initially in the soil liner. Therefore, the initial condition for the SL is expressed as:

$$C_{\mathrm{SL}}(z,0)=0, L_{\mathrm{GCL}}<z<L_{\mathrm{GCL}}+L_{\mathrm{SL}} \tag{7}$$

At the top surface of the GCL, the contaminant concentration is prescribed by the leachate concentration:

$$C_{\mathrm{GCL}}(0,t)=C_0, t>0 \tag{8}$$

The bottom boundary condition is assumed to be semi-infinite, can be expressed as:

$$\frac{\partial c_{\mathrm{SL}}}{\partial z}(L_{\mathrm{GCL}}+L_{\mathrm{SL}},t)=0 \tag{9}$$

The continuity conditions between the GCL and SL layers can be expressed as:

$$C_{GCL}(z=L_{GCL},t)=C_{SL}(z=L_{GCL},t), \tag{10}$$

$$n_{GCL}D_{GCL}\frac{\partial C_{GCL}(z=L_{GCL},t)}{\partial Z}=n_{SL}D_{SL}\frac{\partial C_{SL}(z=L_{GCL},t)}{\partial Z} \tag{11}$$

# 3. PINN Methodology

## 3.1 Neural-network Architecture

In this study, physics-informed neural networks (PINNs) are developed to approximate contaminant transport through the two-layer GCL/SL composite liner system. The schematic of PINN-based models for analyzing contaminant transport in GCL/SL composite liner system is shown in Fig. 2. The model output is expressed in terms of the relative concentration:

$$c_i(z,t)=\frac{C_i(z,t)}{C_0}, i=\mathrm{GCL},\mathrm{SL} \tag{12}$$

where $C_i(z,t)$ is the pore-water contaminant concentration in layer $i$, and $C_0$ is the prescribed contaminant concentration in the overlying leachate. Because the spatial coordinate is normalized separately in each layer, the derivatives in the governing-equation residuals must be scaled using the corresponding layer thickness. The normalized coordinates are:

$$z_g=\frac{z}{L_{GCL}}, 0\le z_g\le 1 \tag{13}$$

$$z_s=\frac{z-L_{\mathrm{GCL}}}{L_{\mathrm{SL}}}, 0\le z_s\le 1 \tag{14}$$

$$\tau = \frac{t}{t_f} \tag{15}$$

where $L_{\mathrm{GCL}}$ and $L_{\mathrm{SL}}$ are the thicknesses of the GCL and SL, respectively, and $t_f$ is the final simulation time.

Because GCL and SL have substantially different thicknesses and material properties, two separate neural networks were adopted. The first neural network approximates the relative concentration field in the GCL, while the second neural network approximates the relative concentration field in the SL. This two-network structure allows the model to handle the strong contrast in material properties between the thin GCL and the much thicker soil liner.

The GCL concentration field is approximated as:

$$\hat{c}_{GCL}(z_g, \tau) = N_{\theta_g}(z_g, \tau; \boldsymbol{\theta_g}) \tag{16}$$

where $\hat{c}_{\mathrm{GCL}}$ is the PINN-predicted relative concentration in the GCL, $N_{\theta_g}$ denotes the GCL neural network, and $\boldsymbol{\theta_g}$ represents its trainable parameters.

Similarly, the SL concentration field is approximated as:

$$\hat{c}_{SL}(z_s, \tau) = N_{\theta_s}(z_s, \tau; \boldsymbol{\theta_s}) \tag{17}$$

where $\hat{c}_{\mathrm{SL}}$ is the predicted relative concentration in the SL, $N_{\theta_s}$ denotes the SL neural network, and $\boldsymbol{\theta_s}$ represents its trainable parameters.

Both neural networks are fully connected feedforward neural networks. For a general network with $M$ hidden layers and $N$ neurons per hidden layer, the mapping can be expressed as:

$$\mathbf{h}^{(0)} = \mathbf{x} \tag{18}$$

$$\mathbf{h}^{(m)} = \sigma\left(\mathbf{W}^{(m)}\mathbf{h}^{(m-1)} + \mathbf{b}^{(m)}\right), m = 1,2, \dots, M \tag{19}$$

$$\hat{c} = \mathbf{W}^{(M+1)}\mathbf{h}^{(M)} + \mathbf{b}^{(M+1)} \tag{20}$$

where $\mathbf{W}^{(m)}$ and $\mathbf{b}^{(m)}$ are the trainable weight matrix and bias vector of the $m$-th layer, respectively, and $\sigma(\cdot)$ is the activation function. Xavier initialization was used for the neural-network weights, and the biases were initialized as zero.

(a)

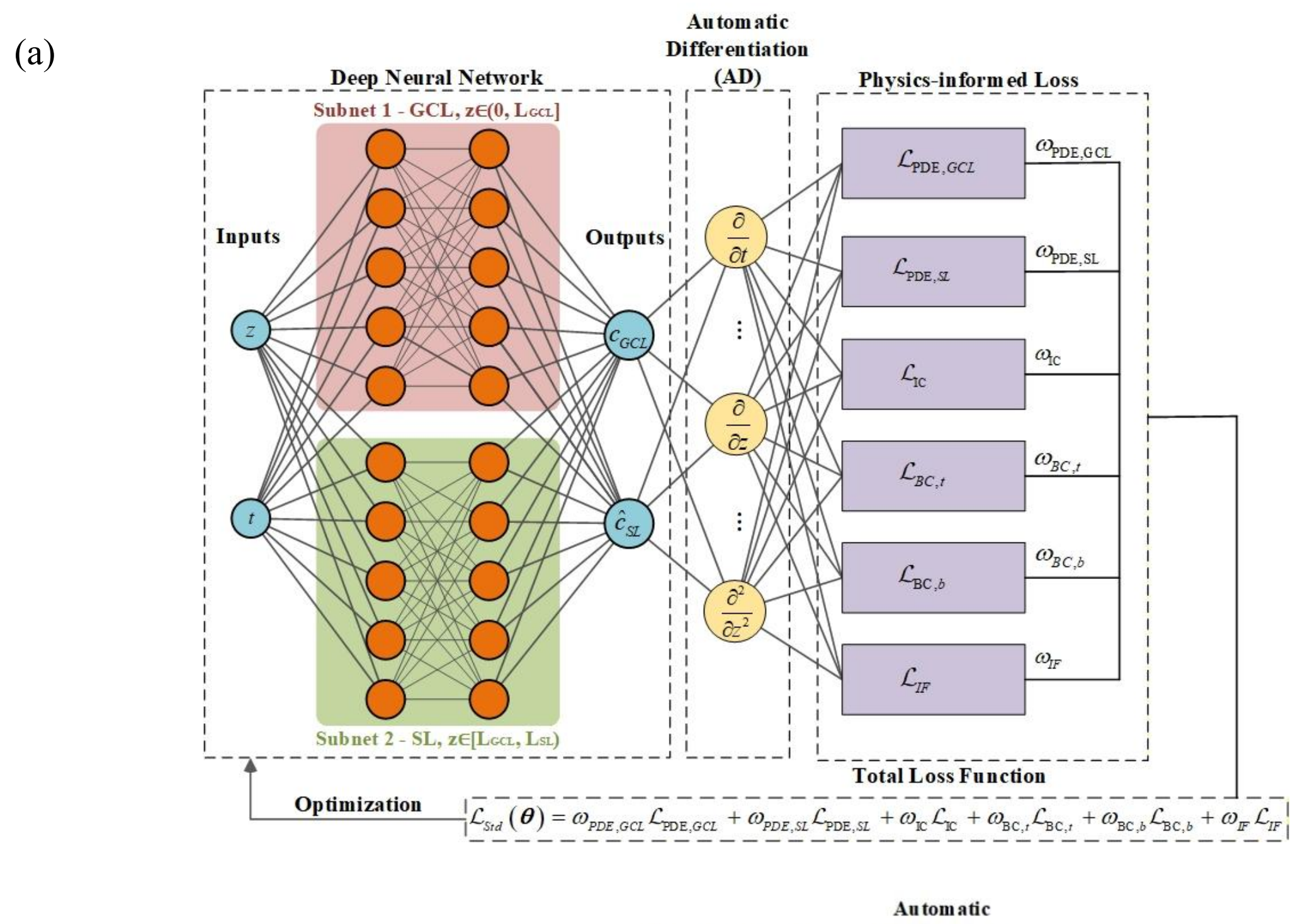

(b)

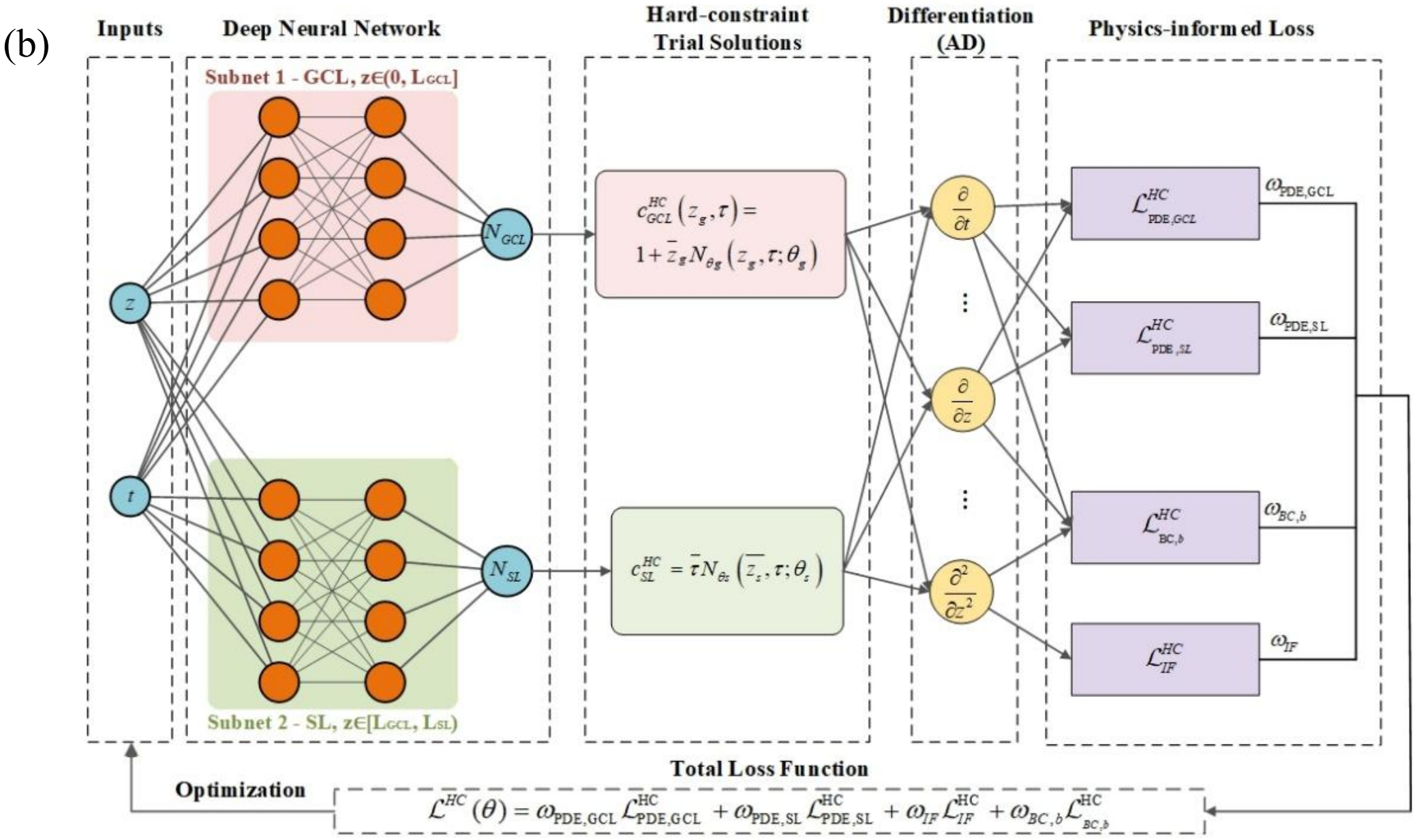

**Fig. 2.** The schematic of PINN-based models (a) Std-PINN, and (b) H-PINN for analyzing contaminant transport in GCL/SL composite liner system

### 3.2 Soft-constrained PINN Formulation

In the soft-constrained PINN, also referred to as the standard PINN or Std-PINN, the raw neural-network outputs are directly used as the approximate concentration fields in the GCL and SL. The governing equations, initial condition, boundary conditions, and GCL/SL interface-continuity conditions are enforced through penalty terms in the total loss function as shown in Fig. 2 (a).

The residual of the steady advection-dispersion-biodegradation equation of GCL layer is defined as:

$$\mathcal{R}_{GCL} = \frac{D_{GCL}}{L_{GCL}^2}\frac{\partial^2 \hat{c}_{GCL}}{\partial z_g^2} + \frac{v_{GCL}}{L_{GCL}}\frac{\partial \hat{c}_{GCL}}{\partial z_g} + \lambda_{GCL} R_{GCL} \hat{c}_{GCL} \tag{21}$$

where $D_{\mathrm{GCL}}$, $v_{\mathrm{GCL}}$, $\lambda_{\mathrm{GCL}}$ are the hydrodynamic dispersion, seepage velocity, and first-order biodegradation coefficient in the GCL, respectively. The GCL residual loss is:

$$L_{PDE,GCL} = \frac{1}{N_{PDE,GCL}} \sum_{k=1}^{N_{PDE,GCL}} \mid \mathcal{R}_{GCL}(z_g^k, \tau^k) \mid^2 \tag{22}$$

where $N_{\mathrm{PDE,GCL}}$ is the number of collocation points sampled in the GCL domain.

Similarly, for the SL layer, the PDE residual is written as:

$$\mathcal{R}_{SL} = \frac{R_{SL}}{t_f}\frac{\partial \hat{c}_{SL}}{\partial \tau} - \frac{D_{SL}}{L_{SL}^2}\frac{\partial^2 \hat{c}_{SL}}{\partial z_s^2} + \frac{v_{SL}}{L_{SL}}\frac{\partial \hat{c}_{SL}}{\partial z_s} + \lambda_{SL} R_{SL} \hat{c}_{SL} \tag{23}$$

where $D_{\mathrm{SL}}$, $v_{\mathrm{SL}}$, $R_{\mathrm{SL}}$, and $\lambda_{\mathrm{SL}}$ are the corresponding transport and degradation parameters in the soil liner. The SL residual loss is:

$$L_{PDE,SL} = \frac{1}{N_{PDE,SL}} \sum_{k=1}^{N_{PDE,SL}} \mid \mathcal{R}_{SL}(z_s^k, \tau^k) \mid^2 \tag{24}$$

At the GCL/SL interface, both concentration continuity and mass-flux continuity are imposed. The concentration-continuity residual is defined as:

$$\mathcal{R}_{IF,c} = \hat{c}_{GCL}(z_g = 1, \tau) - \hat{c}_{SL}(z_s = 0, \tau) \tag{25}$$

and the corresponding loss:

$$L_{IF,c} = \frac{1}{N_{IF}} \sum_{k=1}^{N_{IF}} | \mathcal{R}_{IF,c}(\tau^k) |^2 \tag{26}$$

The mass-flux-continuity residual is expressed as:

$$\mathcal{R}_{IF,J} = \frac{n_{GCL} D_{GCL}}{L_{GCL}} \frac{\partial \hat{c}_{\mathrm{GCL}}}{\partial z} - \frac{n_{SL} D_{SL}}{L_{SL}} \frac{\partial \hat{c}_{\mathrm{SL}}}{\partial z} \tag{27}$$

where $n_{\mathrm{GCL}}$ and $n_{\mathrm{SL}}$ are the porosities of the GCL and SL, respectively. The corresponding flux-continuity loss is:

$$L_{IF,J} = \frac{1}{N_{IF}} \sum_{k=1}^{N_{IF}} | \mathcal{R}_{IF,J}(\tau^k) |^2 \tag{28}$$

Thus, the total interface-continuity loss is:

$$L_{IF} = L_{IF,c} + L_{IF,J} \tag{29}$$

The boundary conditions loss at the top of the GCL is imposed as

$$L_{BC,t} = \frac{1}{N_{BC,t}} \sum_{k=1}^{N_{BC,t}} | \hat{c}_{GCL}(z_g = 0, \tau^k) - 1 |^2 \tag{30}$$

The bottom boundary condition of the SL is imposed as:

$$L_{BC,b} = \frac{1}{N_{BC,b}} \sum_{k=1}^{N_{BC,b}} | \frac{\partial \hat{c}_{SL}}{\partial z_s} (z_s = 1, \tau^k) |^2 \tag{31}$$

The initial-condition loss of  SL is imposed as

$$\mathcal{L}_{\mathrm{IC}} = \frac{1}{N_{IC,SL}} \sum_{i=1}^{N_{IC,SL}} [\hat{c}_{SL}(z_i, 0)]^2 \tag{32}$$

The total loss function of the Std-PINN is therefore

$$\mathcal{L}_{Std} = w_{GCL} \mathcal{L}_{PDE,GCL} + w_{SL} \mathcal{L}_{PDE,SL} + w_{IF} \mathcal{L}_{IF} + w_t \mathcal{L}_{BC,t} + w_b \mathcal{L}_{BC,b} + w_{IC} \mathcal{L}_{IC} \tag{33}$$

where $w_{\mathrm{GCL}}$, $w_{\mathrm{SL}}$, $w_{\mathrm{IF}}$, $w_t$, $w_b$, and $w_{\mathrm{IC}}$ are weighting factors used to balance the different loss components.

### 3.3 H-PINN Formulation

As shown in Fig. 2(b), in the H-PINN, selected physical constraints are embedded directly into the neural-network trial functions instead of being imposed only through penalty terms in the loss function. This formulation reduces the optimization burden associated with simultaneously minimizing multiple competing loss components. In this study, the prescribed concentration boundary at the top of the GCL and the initial condition in the SL are imposed exactly through output transformations.

For the GCL, the hard-constrained solution is defined as

$$\hat{c}_{\mathrm{GCL}}^{\mathrm{HC}}(z_g,\tau) = 1 + \bar{z}_g N_{\theta_g}(z_g,\tau;\boldsymbol{\theta_g}) \tag{34}$$

When $\bar{z}_g = 0$ at the top of the GCL, Eq. (34) gives:

$$\hat{c}_{\mathrm{GCL}}^{\mathrm{HC}}(0,\tau) = 1 \tag{35}$$

Therefore, the prescribed source concentration at the GCL/leachate boundary is satisfied exactly for all training iterations. No separate top-boundary loss term is required in the hard-constrained formulation.

For the SL, the hard-constrained trial solution is constructed as:

$$\hat{c}_{\mathrm{SL}}^{\mathrm{HC}} = \bar{\tau} N_{\theta_s}(\bar{z}_s,\tau;\boldsymbol{\theta_s}) \tag{36}$$

At the initial time, where $\tau = 0$,

$$\hat{c}_{\mathrm{SL}}^{\mathrm{HC}}(z,0) = 0 \tag{37}$$

Thus, the initial contaminant-free condition in the SL is also satisfied exactly. No separate SL initial-condition loss term is required in the hard-constrained formulation.

The remaining physical constraints, including the GCL governing-equation residual, the SL governing-equation residual, and the GCL/SL interface-continuity conditions, are still enforced through the loss function. The interface concentration and flux conditions are retained as soft constraints because they couple the two neural networks and depend on the material properties of both layers. Therefore, the total loss function for the H-PINN is written as:

$$\mathcal{L}_{HC} = w_{GCL}\mathcal{L}_{PDE,GCL}^{HC} + w_{SL}\mathcal{L}_{PDE,SL}^{HC} + w_{IF}\mathcal{L}_{IF}^{HC} + w_b\mathcal{L}_{BC,b}^{HC} \tag{38}$$

where the superscript HC indicates that the residuals are evaluated using the hard-constrained trial solutions.

## 3.4 Inverse Modeling Formulation

In addition to forward prediction, the proposed PINN framework was extended to inverse modeling to estimate uncertain degradation parameters from limited concentration observations. In practical GCL/SL composite liner systems, biodegradation parameters are often difficult to determine directly because they depend on contaminant type, microbial activity, moisture condition, temperature, and geochemical environment. Therefore, the SL biodegradation half-life, $t_{1/2,SL}$, was selected as the target inverse parameter in this study.

In the inverse problem, the neural-network weights, biases, and selected physical parameters are optimized simultaneously. The observation dataset is denoted as:

$$\mathcal{D}_{obs} = \{z_j, t_j, c_j^{obs}\}_{j=1}^{N_{obs}} \tag{39}$$

where $z_j$ and $t_j$ are the observation location and time, respectively, $c_j^{obs}$ is the observed relative concentration, and $N_{obs}$ is the total number of observation points. In this study, the observation data were generated from the reference solution at selected locations within the SL layer and at the bottom of the liner system. These data represent limited monitoring information that may be available in practical contaminant-transport assessment.

The data loss is defined as:

$$\mathcal{L}_{data} = \frac{1}{N_{obs}} \sum_{j=1}^{N_{obs}} \left[\hat{c}_{PINN}\,(z_j, t_j) - c_j^{obs}\right]^2 \tag{40}$$

where $\hat{c}_{PINN}\,(z_j, t_j)$ is the predicted relative concentration at the observation point. For observation points located in the SL layer, the prediction is evaluated using the SL neural-network prediction.

For the H- PINN, the inverse loss function is written as

$$L_{HC} = w_{GCL} L_{PDE,GCL}^{HC} + w_{SL} L_{PDE,SL}^{HC} + w_{IF} L_{IF}^{HC} + w_b L_{BC,b}^{HC} + w_d \mathcal{L}_{data} \tag{41}$$

where $\mathcal{L}_{data}$, $w_{data}$ are the concentration observation loss, and weighting factors respectively.

For the noisy-observation cases, synthetic noise was added to the reference concentration data as:

$$c_{obs}^{noise} = c_{obs}(1+\delta\varepsilon) \tag{42}$$

where $\delta$ is the prescribed relative noise level and $\varepsilon$ is a random variable sampled from a standard normal distribution. The noise levels considered in this study were 0%, 10%, 20%, 30%, and 50%.

# 4. Model Implementation

For the proposed Std-PINN and H-PINN models, two independent fully connected neural networks were used to approximate the relative concentration fields in the GCL and SL, respectively. Unless otherwise stated, each neural network contained four hidden layers with 128 neurons per hidden layer. All neural network training was executed in Python 3.12.3 with PyTorch 2.5.0, on an NVIDIA GeForce RTX 4090. For the forward simulations, the GCL domain was discretized from $z = 0$ to 0.0138 m using 101 uniformly distributed collocation points. The SL domain was discretized from $z = 0.0138$ to 5.0138 m and from $t$ =0 to 100 years using a structured 401 × 401 space-time grid. The spatial and temporal inputs were normalized before being passed into the neural networks to improve training stability.

The neural-network weights were initialized using Xavier initialization, and the biases were initialized as zero. The Tanh activation function was adopted for both Std-PINN and H-PINN because it is smooth and differentiable, making it suitable for automatic differentiation of the first- and second-order derivatives required in the governing transport equations. All derivatives in the PDE residuals and interface-continuity terms were computed using PyTorch automatic differentiation. The models were trained using a two-stage optimization strategy. Adam was first used to obtain a stable initial solution, followed by L-BFGS optimization to further reduce the physics-based residuals. The Std-PINN and H-PINN models were trained using 20,000 Adam iterations followed by L-BFGS optimization.

## 4.1 Model Evaluation Indices

The analytical solution of Guan et al. (2014) was used for breakthrough-curve validation, whereas the COMSOL finite-element solution was used to generate full spatiotemporal concentration fields for contour-level error evaluation. Unless otherwise noted, all Std-PINN and H-PINN results are compared against these FEM and analytical solutions.

For the GCL/SL composite liner system, error indices can be calculated over the entire computational domain or separately within the GCL and SL layers. The prediction accuracy of the PINN models was evaluated using the mean absolute error (MAE) and mean relative error (MRE). The MAE measures the average absolute difference between the PINN prediction and the reference solution, while the MRE quantifies the relative prediction error in percentage form.

$$MAE = \frac{1}{N}\sum_{i=1}^{N} | c_i^{PINN} - c_i^{ref} | \tag{43}$$

$$MRE = \frac{1}{N}\sum_{i=1}^{N} \frac{| c_i^{PINN} - c_i^{ref} |}{| c_i^{ref} | + \varepsilon} * 100\% \tag{44}$$

where $c_i^{\mathrm{PINN}}$ and $c_i^{\mathrm{ref}}$ are the predicted and reference normalized concentrations at the $i$-th evaluation point, respectively; $N$ is the total number of evaluation points; and $\varepsilon$ is a small constant used to avoid division by zero when the reference concentration approaches zero.

# 5. Results and Discussion

This section presents the results with reference to the analytical solution for one-dimensional solute transport through a GCL/SL composite liner considering biodegradation, as reported by Guan et al. (2014). The liner system consists of a 0.0138 m thick GCL overlying a 5.0 m thick soil liner. Benzene was selected as the representative organic contaminant, and the initial contaminant concentration in the leachate was taken as $C_0$ =1 mg/L. The porosities of the GCL and SL are 0.86 and 0.40, respectively, and the retardation factor of the SL is $R_d = 2.134$. In the PINN simulations, the diffusion coefficient of the GCL and SL are 3.6 $\times 10^{-10}$ and .8.9 $\times 10^{-10}$ m²/s, respectively. The first-order biodegradation coefficient is .2.2 $\times$ $10^{-9}$ $s^{-1}$ corresponding to a contaminant half-life of 10 years. Three leachate heads, $H_w$ = 0.3, 3.0, and 10.0 m, were considered to evaluate the effect of leachate head on breakthrough behavior of GCL/SL composite liner. The seepage velocities used in the GCL and SL layers under different leachate-head conditions were determined according to the calculation procedure reported by Guan et al. (2014). All Std-PINN and H-PINN predictions are compared against the analytical solution of Guan et al. (2014) and the numerical reference solution used for verification. The parameters used in the GCL/SL contaminant-transport model are shown in Table 1.

**Table 1** Material properties of the GCL/SL composite liner

| Parameter | Symbol | GCL | SL | Unit |
|---|---|---|---|---|
| Layer thickness | L | 0.0138 | 5.0 | m |
| Porosity | n | 0.86 | 0.40 | - |
| Diffusion coefficient | $D^*$ | $3.6 \times 10^{-10}$ | $8.9 \times 10^{-10}$ | $m^2/s$ |
| Retardation factor | $R_d$ | - | 2.134 | - |
| Dry density | $\rho_b$ | - | 1.62 | $g/cm^3$ |
| Distribution coefficient | $K_d$ | - | 0.28 | mL/g |
| First-order degradation coefficient | $\lambda$ | $2.20 \times 10^{-9}$ | $2.20 \times 10^{-9}$ | $s^{-1}$ |
| Corresponding half-life | $t_{1/2}$ | 10 | 10 | year |
| Initial leachate concentration | $C_0$ | 1.0 | 1.0 | mg/L |
| Simulation duration | $t_f$ | 100 | 100 | year |

## 5.1 Training Behavior of Std-PINN and H-PINN

Fig. 3 shows the evolution of individual loss components during training for the Std-PINN and H-PINN under the representative leachate head condition of $H_w = 0.3$ m. Both models exhibit a rapid decrease in total loss during the early stage of Adam optimization, indicating that the neural networks quickly learn the dominant features of the concentration field and the associated physical constraints. After the initial reduction, the losses enter a slower convergence stage, where small oscillations and occasional spikes are observed, particularly in the PDE residual. These spikes are expected in PINN training because the PDE loss involves higher-order automatic differentiation and is more sensitive to local changes in the network parameters.

For the Std-PINN, the total loss decreases rapidly at the beginning and then gradually stabilizes during the remaining Adam iterations. The IC, PDE, and IF losses are reduced to relatively low levels, while the BC loss remains one of the dominant components of the total loss. This indicates that the Std-PINN can satisfy the governing equation and interface continuity reasonably well, but the enforcement of boundary conditions through soft penalty terms remains comparatively more difficult. The subsequent L-BFGS optimization provides further refinement, although the improvement is moderate because the main reduction has already been achieved during the Adam stage.

For the H-PINN, the initial condition is embedded directly into the trial solution; therefore, no separate IC loss term is required. The PDE, BC, and IF losses generally decrease during training, and the total loss remains stable after the early rapid reduction. Compared with the Std-PINN, the H-PINN shows a clearer separation of loss components, with the boundary-condition loss contributing significantly to the total loss during most of the training process. The interface loss remains relatively small, suggesting that the coupling between the GCL and SL domains is effectively maintained. Although the PDE loss exhibits several sharp fluctuations during Adam optimization, it remains at a relatively low magnitude after convergence. The L-BFGS stage further reduces the total loss and smooths the final convergence behavior. This behavior supports the advantage of the hard-constrained formulation in reducing the number of competing loss terms and improving the physical consistency of the learned solution.

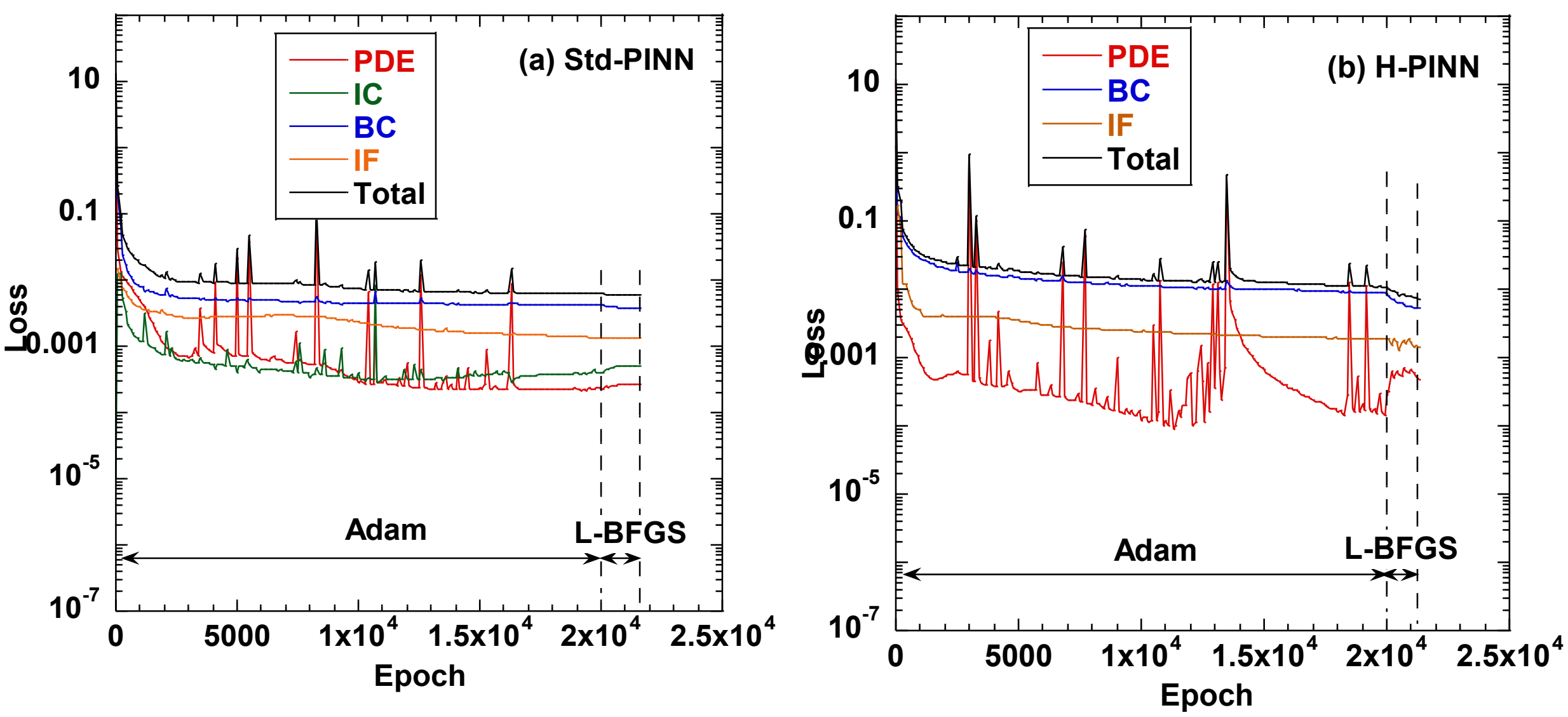

**Fig. 3.** Evolution of loss components during training: (a) Std-PINN and (b) H-PINN with $H_w = 0.3$ m

## 5.2 Comparison with Reference Solution

Breakthrough time is defined as the time required for the concentration of a solute, e.g., benzene, to reach the breakthrough concentration at the bottom of the liner system (Acar and Haider, 1990) in this case. The breakthrough concentration for benzene is 0.005 mg/L on the basis of the drinking water standard of (USEPA, 2001). Fig. 4 compares the breakthrough curves at the bottom of the soil liner under different leachate heads. The relative concentration $C/C_0$ increases with time for all cases and eventually approaches a quasi-steady value. As the leachate head increases from $H_w = 0.3$ m to $H_w = 10.0$ m, the breakthrough shifts upward and to the left, indicating that a larger leachate head accelerates advective contaminant transport and results in earlier breakthrough. The final relative concentration also increases with leachate head, from approximately 0.55 for $H_w = 0.3$ m to about 0.67 for $H_w = 3.0$ m and 0.80 for $H_w = 10.0$ m. This trend confirms that leachate head is a key factor controlling the long-term containment performance of the GCL/SL composite liner system. The H-PINN predictions show closer agreement with both the FEM results and the analytical solution for all three leachate-head cases. In comparison, the Std-PINN captures the general breakthrough trend but shows larger deviations, especially for the higher leachate-head cases. This indicates that the hard-constrained formulation improves the prediction of transient contaminant breakthrough by better satisfying the imposed physical constraints.

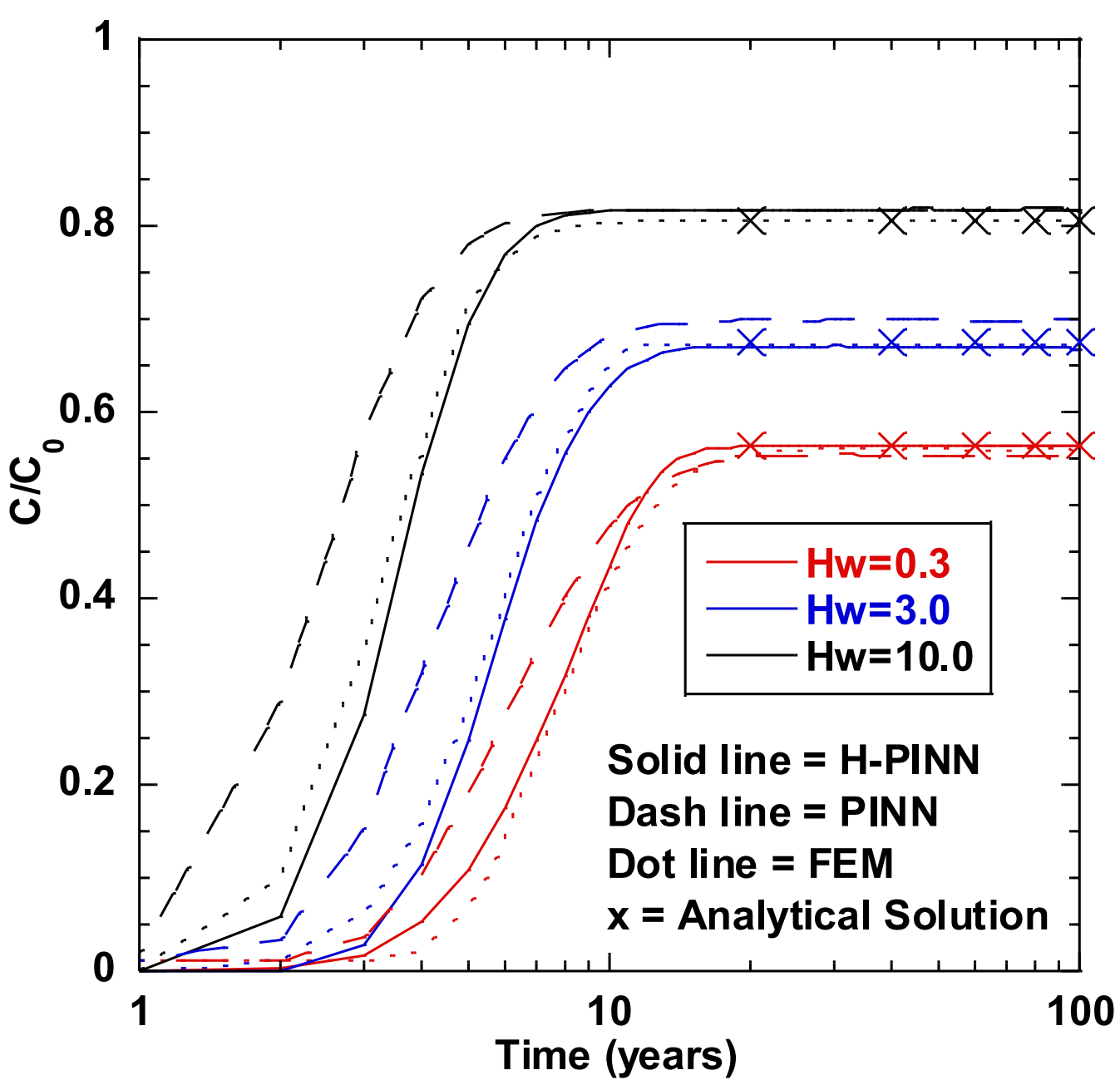

**Fig. 4.** Breakthrough curves at the bottom of the soil liner under different leachate heads.

Fig. 5 compares the FEM reference solution, Std-PINN prediction, and corresponding absolute error for the relative concentration field $C/C_0$ under different leachate heads. Overall, the Std-PINN captures the main spatiotemporal migration pattern of contaminant transport. For all three leachate heads, the concentration remains relatively high near the GCL/leachate interface and gradually propagates downward into the SL with time. As the leachate head increases from $H_w = 0.3$ m to $H_w = 10.0$ m, the contaminant front advances more rapidly, indicating that a larger hydraulic gradient accelerates advective transport through the liner system. The FEM results show a clear transition zone between the high-concentration region near the top of GCL and the low-concentration region near the bottom of SL , and this transition becomes steeper under larger leachate heads.

Although the Std-PINN reproduces the overall concentration distribution, noticeable discrepancies are observed during the early to intermediate transport period. The absolute-error contours indicate that the largest errors are mainly concentrated around the steep-gradient region rather than in the fully contaminated or nearly uncontaminated zones. This behavior suggests that the Std-PINN can approximate the smooth regions of the solution but has difficulty accurately resolving the sharp concentration transition caused by coupled advection-diffusion transport. The error magnitude also increases with leachate head, with the largest localized error occurring for $H_w = 10.0$ m, where the concentration front develops more rapidly and the transport process becomes more advection dominated.

Fig. 6 presents the corresponding results obtained using the H-PINN model. Similar to the Std-PINN, the H-PINN successfully captures the acceleration of solute transport with increasing leachate head. However, compared with Fig. 5, the H-PINN produces a closer agreement with the FEM reference solution, particularly near the concentration transition front. The absolute-error contours show that the error is more localized and generally reduced, especially at later times after the concentration profile becomes smoother. This improvement indicates that the hard-constrained formulation enhances physical consistency by embedding prescribed constraints directly into the neural-network approximation, thereby reducing the burden on the loss-function optimization.

For the low and moderate leachate-head cases, the H-PINN maintains relatively small errors over most of the computational domain, with localized discrepancies mainly appearing near the early-time concentration front. For the high leachate-head case, some errors remain near the soil liner during the early stage, but the error decreases substantially as time progresses. These results demonstrate that the H-PINN provides a more stable prediction than the Std-PINN under different leachate heads, particularly when the transport process involves strong concentration gradients and rapid contaminant migration.

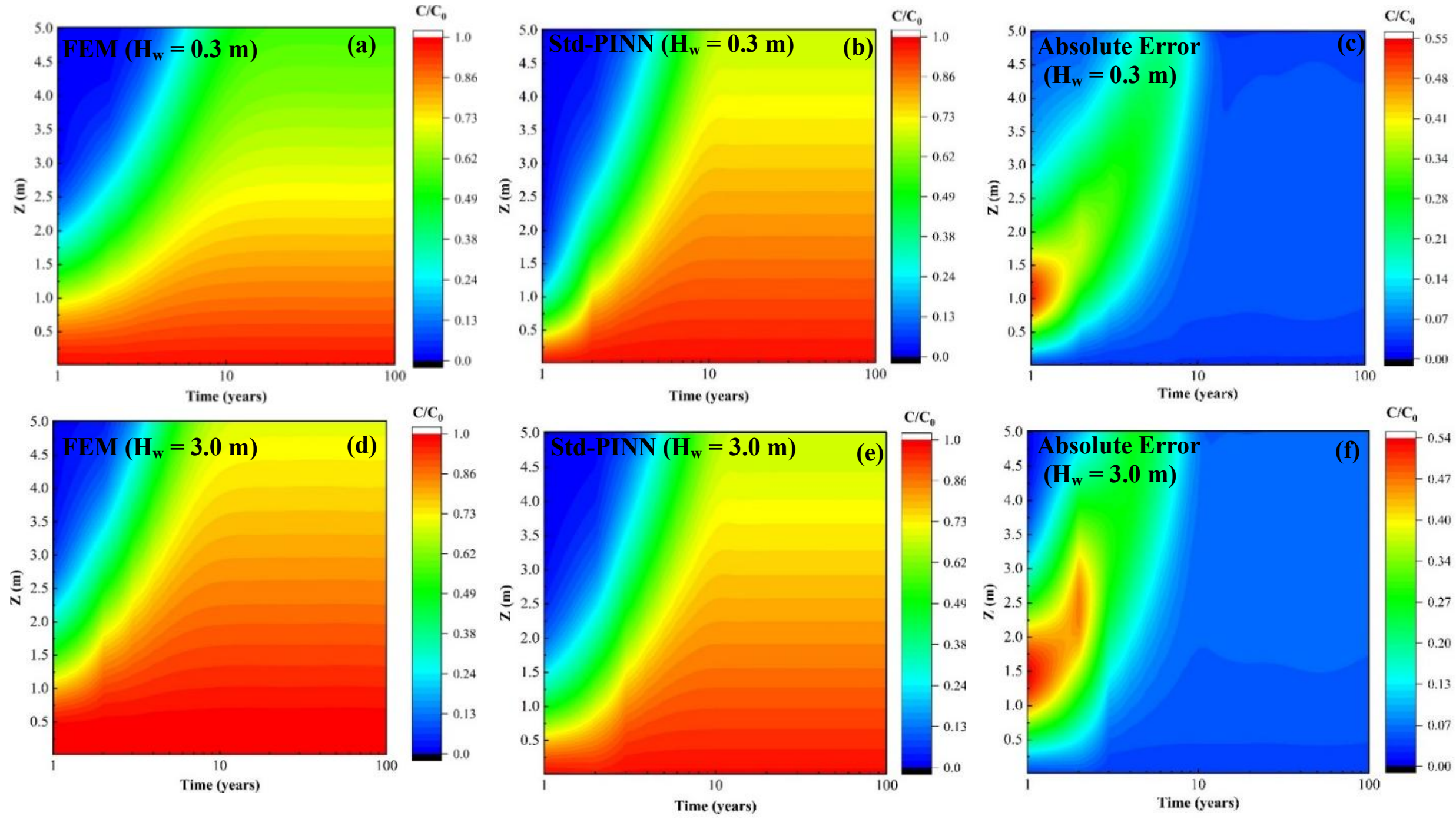

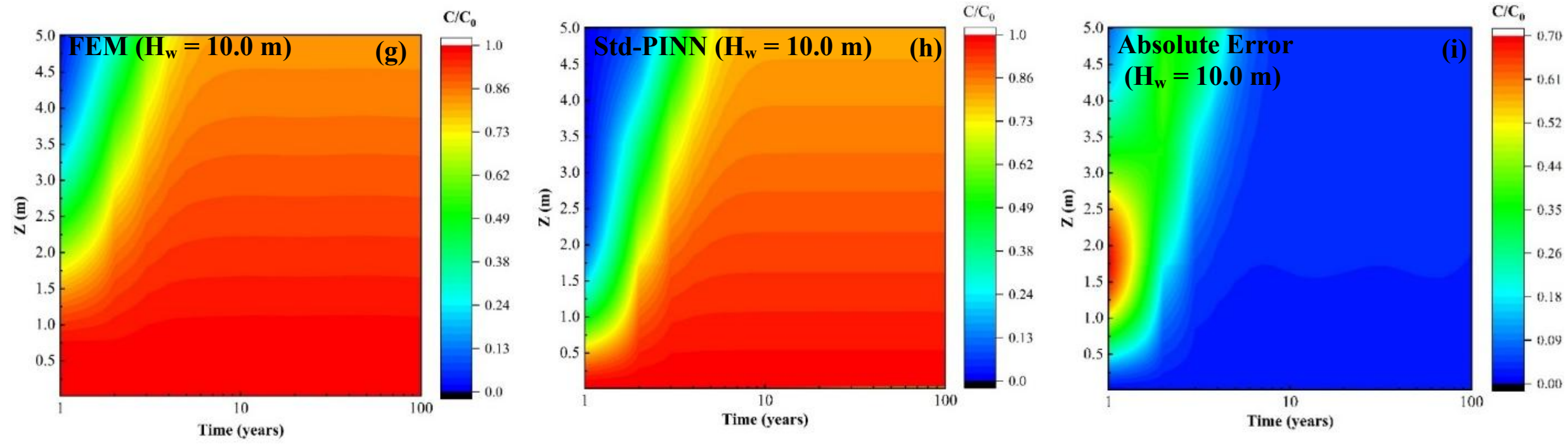

**Fig. 5.** Spatiotemporal distributions of relative concentration $C/C_0$ and absolute error for the Std-PINN under different leachate heads: (a-c) $H_w = 0.3$ m, (d-f) $H_w = 3.0$ m, and (g-i) $H_w = 10.0$ m.

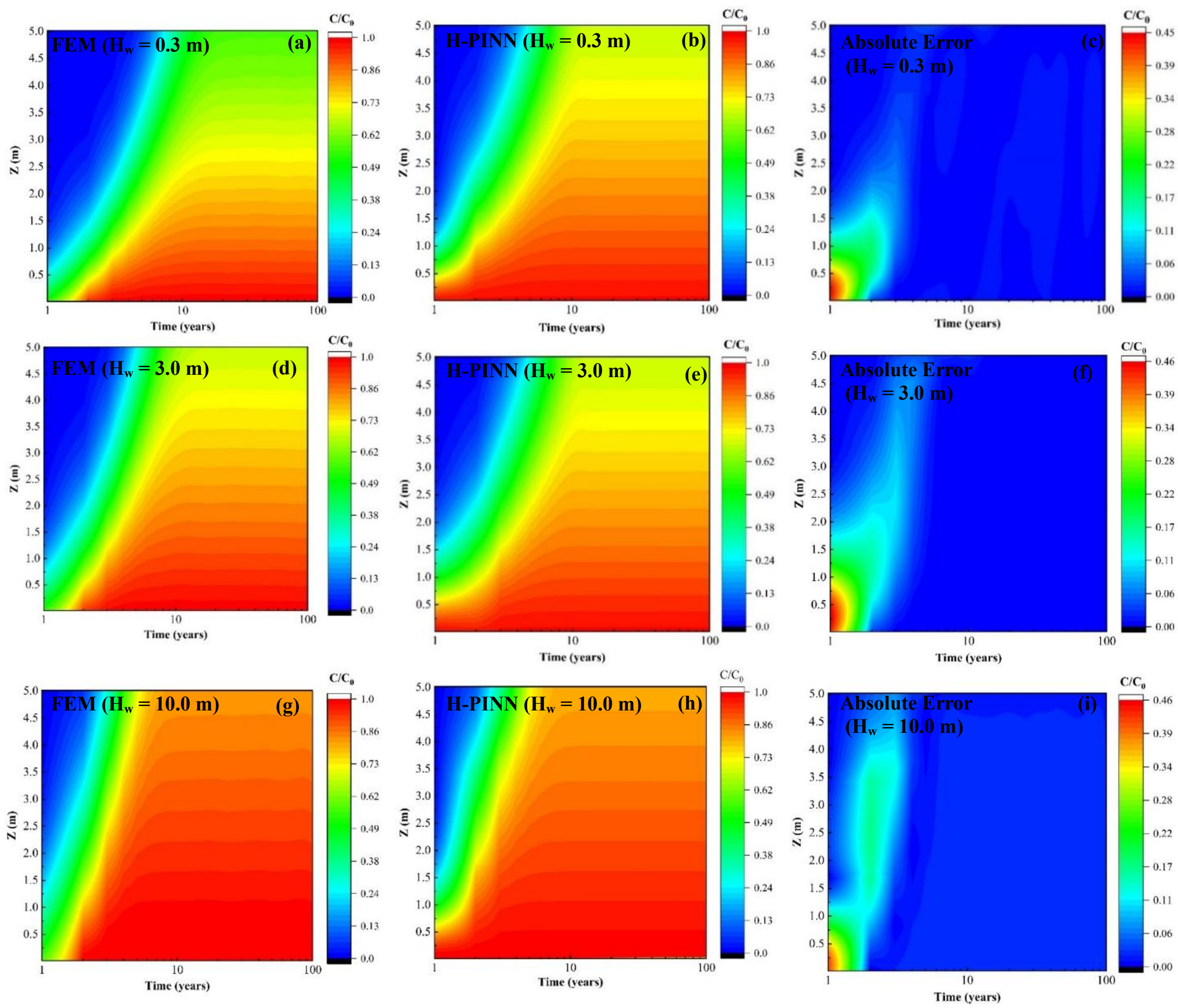

**Fig. 6.** Spatiotemporal distributions of relative concentration $C/C_0$ and absolute error for the H-PINN under different leachate heads: (a-c) $H_w = 0.3$ m, (d-f) $H_w = 3.0$ m, and (g-i) $H_w = 10.0$ m.

Fig. 7 quantitatively compares the prediction accuracy of the Std-PINN and H-PINN models under different leachate heads using MAE and MRE. Overall, the H-PINN produces substantially lower errors than the Std-PINN for all leachate heads, confirming the improved accuracy observed in the contour plots. For the Std-PINN, the MAE values are 0.058, 0.066, and 0.067 for leachate heads of 0.3, 3.0, and 10.0 m, respectively, while the corresponding MRE values are 9.10%, 18.18%, and 19.16% as shown in Table 2. The relatively large MRE values indicate that the Std-PINN has difficulty accurately resolving the concentration field, particularly during the early transport stage when the concentration front develops rapidly and the solution gradient is steep. In contrast, the H-PINN significantly reduces both MAE and MRE. The MAE values decreased to 0.011, 0.011, and 0.023 for leachate heads of 0.3, 3.0, and 10.0 m, respectively. Similarly, the MRE values of the H-PINN remain very small, generally below 3.2 % for all cases. This demonstrates that embedding the physical constraints directly into the neural-network approximation improves both global prediction accuracy and numerical stability.

The improvement is especially evident for $H_w = 3.0$ m, where the Std-PINN shows the largest MRE of 18.18%, while the H-PINN reduces the MRE to 2.1%. Although the H-PINN error slightly increases under the highest leachate head, this is expected because the larger hydraulic gradient accelerates contaminant migration and produces a sharper concentration front. Nevertheless, the H-PINN still outperforms the Std-PINN by a large margin, indicating that the hard-constrained formulation is more robust for contaminant transport problems involving strong advective effects and rapidly evolving concentration gradients.

**Table 2** Quantitative comparison of Std-PINN and H-PINN prediction accuracy

| Leachate head, $H_w$ (m) | Std-PINN | | H-PINN | |
|---|---|---|---|---|
| | MAE | MRE (%) | MAE | MRE (%) |
| 0.3 | 0.058 | 9.10 | 0.011 | 2.20 |
| 3.0 | 0.066 | 18.18 | 0.011 | 2.08 |
| 10.0 | 0.067 | 19.16 | 0.023 | 3.14 |

Note: MAE = mean absolute error; MRE = mean relative error.

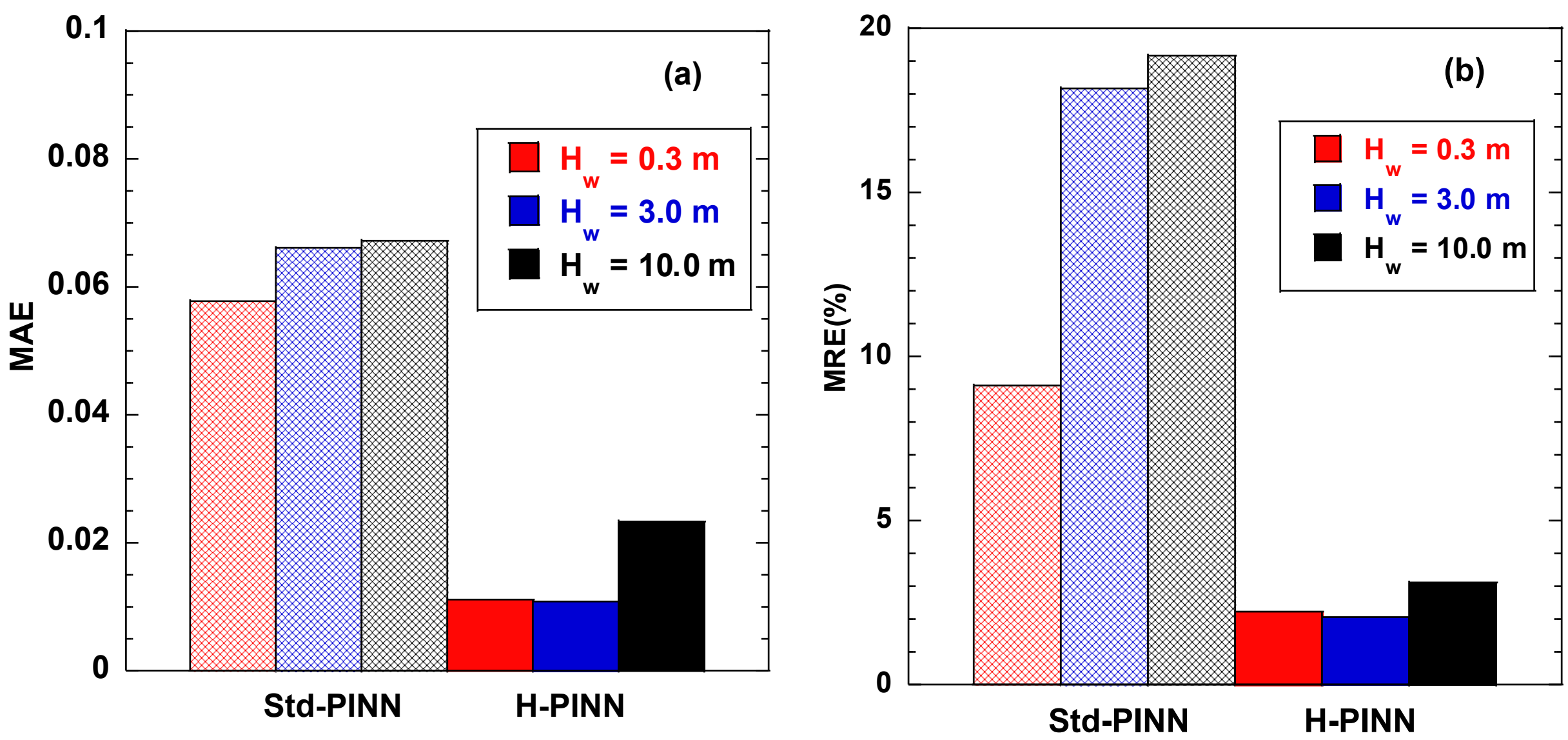

**Fig. 7.** Quantitative evaluation of model accuracy for Std-PINN and H-PINN under different leachate-head conditions: (a) MAE and (b) MRE.

## 5.3 Parameter Study

### 5.3.1 Effect of activation functions

The effect of activation functions on the performance of the H-PINN model was evaluated in this study. The ReLU, Sigmoid, ELU, and Tanh functions are four of the main nonlinear activation functions used in deep learning (Nwankpa et al., 2018). Fig. 8 compares the predicted normalized concentration histories at the bottom of the SL layer and the corresponding MRE values obtained using different activation functions. As shown in Fig. 8(a), the Tanh activation function provides the closest agreement with the analytical solution. ELU and Sigmoid also capture the overall concentration evolution, but with relatively larger deviations. In contrast, the ReLU activation function shows a delayed concentration response and noticeable oscillations near the rapid concentration increase stage. This indicates that Tanh is suitable for representing the smooth concentration field in the GCL/SL contaminant transport problem. The quantitative comparison in Fig. 8(b) further confirms this observation. The Tanh activation function gives the lowest MRE, followed by ELU, Sigmoid, and ReLU. Therefore, Tanh was selected as the activation function for the subsequent PINN simulations in this study.

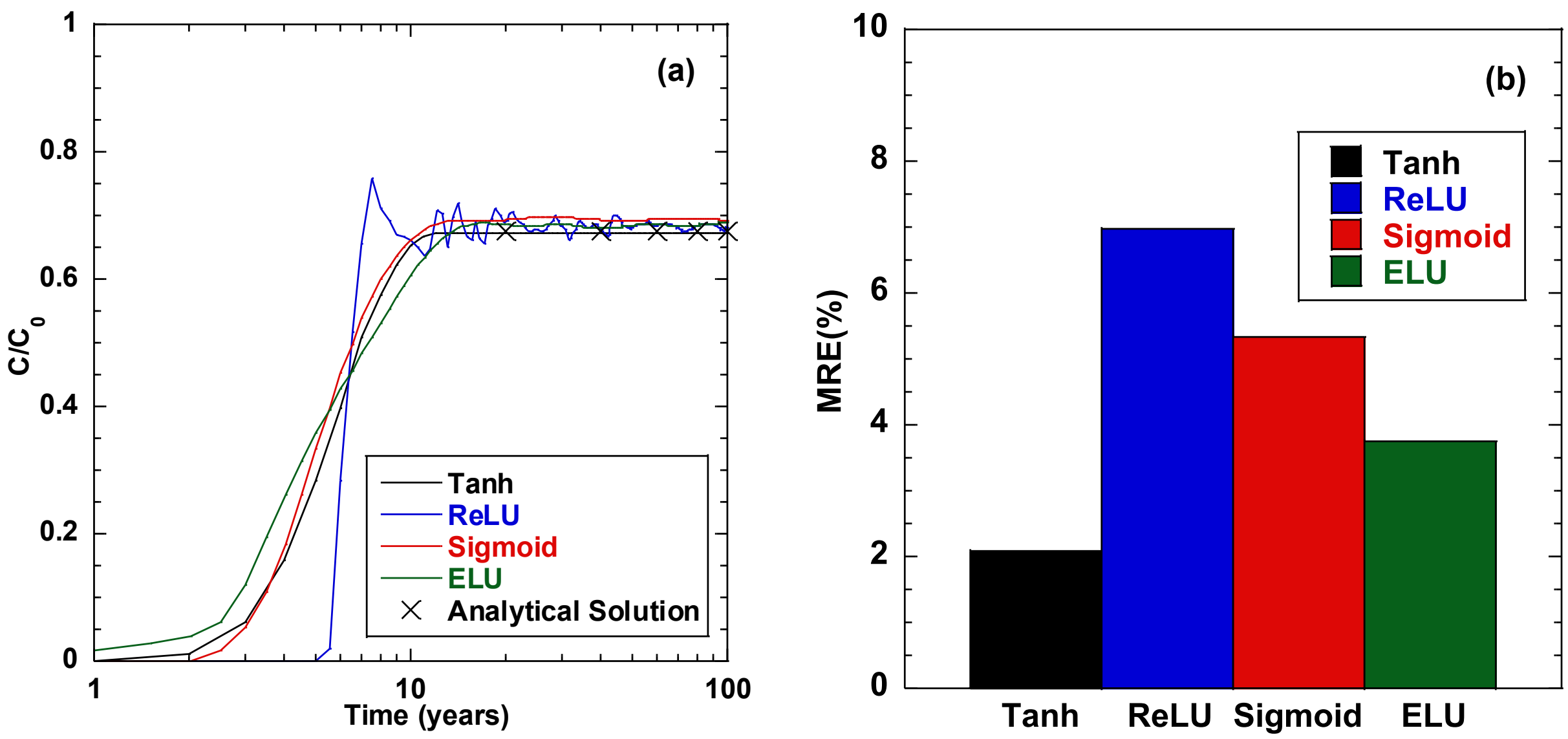

**Fig. 8.** Effect of activation functions on PINN prediction accuracy: (a) normalized concentration history at the bottom of the SL layer; and (b) comparison of MRE values for different activation functions with $H_w$ = 3.0 m

### 5.3.2 Effect of neural-network structure

Fig. 9 shows the influence of neural-network structure on the prediction accuracy of the H-PINN model for the $H_w = 3.0$ m. The number of hidden layers varied from 2 to 5, and the number of neurons per hidden layer varied from 32 to 256. All other model settings, including the activation function, collocation-point allocation, loss weights, and training procedure, were kept unchanged. The results indicate that the network structure has a noticeable effect on prediction accuracy. When a relatively small network is used, such as 2 hidden layers with 32 neurons per layer, the model gives the largest error, suggesting that the network capacity is insufficient to accurately approximate the nonlinear concentration field in the GCL/SL composite liner system. Increasing the network depth from 2 to 4 hidden layers generally reduces the MAE, indicating that additional hidden layers improve the representational capability of the H-PINN model.

However, further increasing the depth to 5 hidden layers does not lead to significant improvement. When the network depth increases from 4 to 5 hidden layers, the reduction in MAE becomes negligible. Similarly, increasing the network width from 128 to 256 neurons provides only marginal improvement in some cases and may slightly increase the error in others. This suggests that the prediction accuracy reaches a plateau once sufficient network capacity is achieved, and that an excessively large network may not improve model performance due to increased optimization difficulty and computational cost. Among the tested structures, the H-PINN model with 4 hidden layers and 128 neurons per hidden layer gives the lowest MAE, with a value of 0.0107.

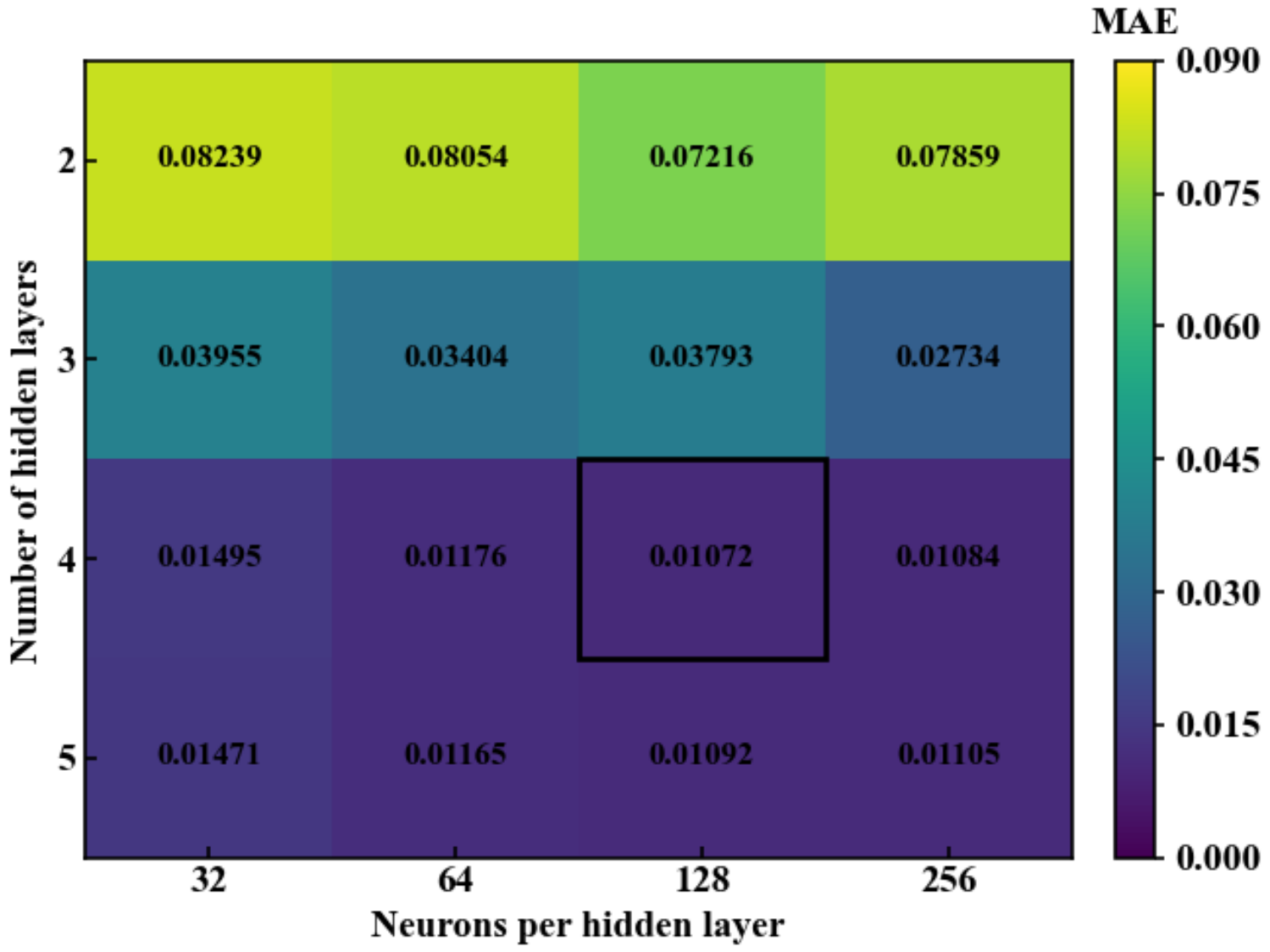

**Fig. 9.** Effect of neural-network structure on the prediction accuracy of the H-PINN model.

### 5.3.3 Effect of SL biodegradation

Fig. 10 illustrates the influence of degradation in the SL on the solute breakthrough behavior of the GCL/SL composite liner system with $H_w = 3.0$ m. The degradation effect was represented by the contaminant half-life in the SL, with half-life values of 10, 30, and 100 years.

As shown in Fig. 10, the H-PINN model accurately reproduces the FEM breakthrough curves for all three half-life values. The breakthrough curves exhibit a clear dependence on the degradation rate in the SL. When the half-life is relatively short, such as $t_{1/2}$ = 10 years, stronger degradation occurs during contaminant migration through the SL, resulting in a lower normalized concentration at the bottom boundary. In contrast, increasing the half-life to 30 and 100 years weakens the degradation effect, leading to progressively higher values of ($C/C_0$). This trend is well captured by the H-PINN model, indicating that the degradation term in the governing equation is effectively incorporated into the physics-informed learning framework.

The comparison between the H-PINN and FEM results further demonstrates that the proposed model can capture both the timing and magnitude of solute breakthrough under different degradation conditions. Minor deviations are mainly observed near the steep transition region of the breakthrough curves, where the concentration changes rapidly with time.

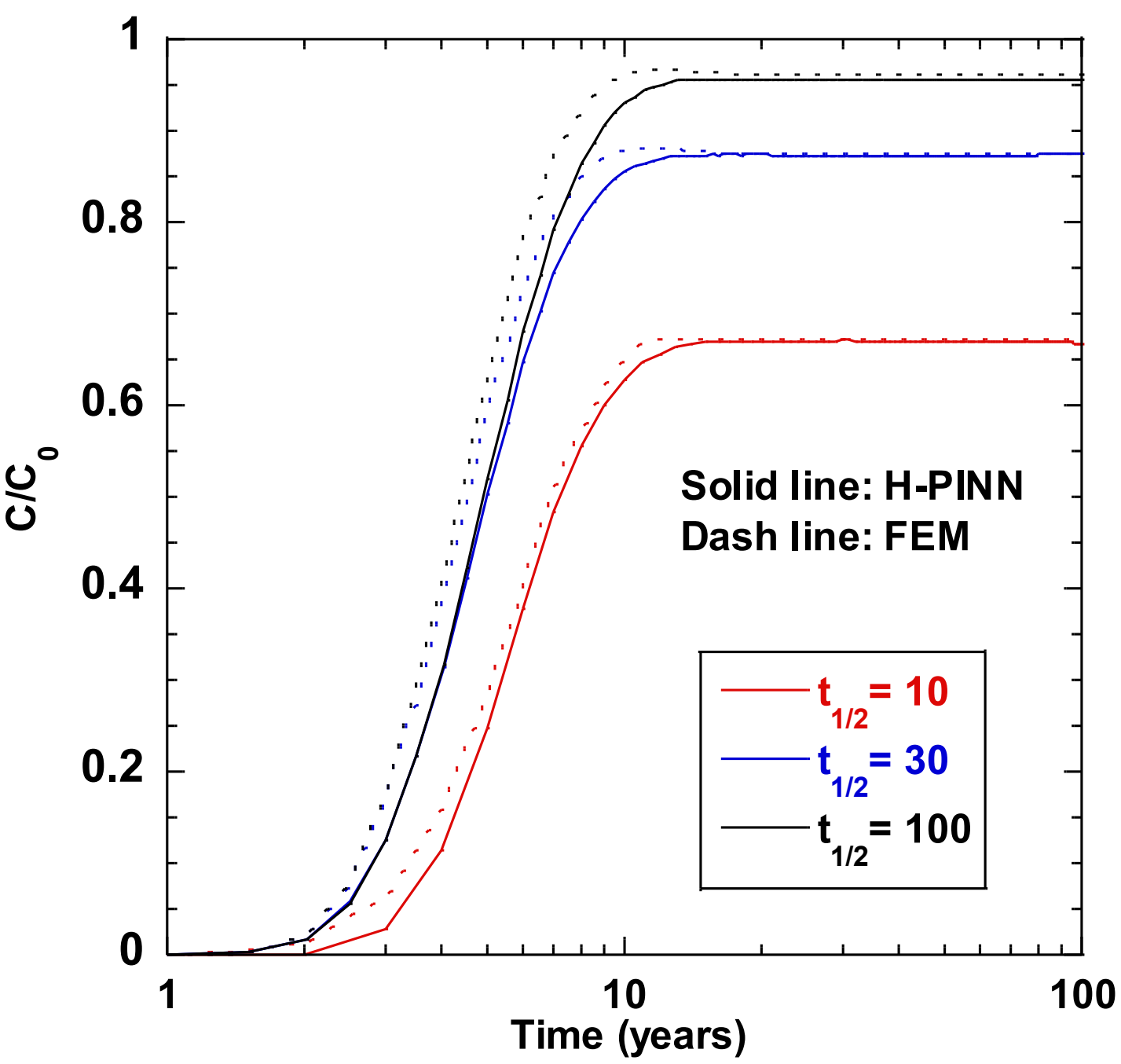

**Fig. 10.** Effect of SL degradation on solute breakthrough curves of the GCL/SL system with $H_w = 3.0$ m.

## 5.4 Inverse Identification of Soil Liner Degradation Half-Life

To further evaluate the capability of the proposed H-PINN framework for parameter identification, inverse simulations were conducted to estimate the degradation half-life of the soil liner. Fig. 11 shows the convergence histories of the identified SL half-life under $H_w = 3.0$ m, with prescribed exact half-life values of 10, 30, and 100 years. For the case with an exact half-life of 10 years, the estimated $t_{1/2}$ rapidly approaches the target value after a short initial fluctuation and remains close to the exact solution during the later training stage. When the exact half-life is 30 years, the estimated $t_{1/2}$ initially overshoots the target value, reaching a higher intermediate value before gradually decreasing and stabilizing near the prescribed value. A similar but more pronounced overshooting behavior is observed for the case with an exact half-life of 100 years, where the inferred $t_{1/2}$ shows a larger transient deviation before converging toward the target value. These results indicate that the proposed inverse H-PINN can recover degradation-related transport parameters over a broad range of half-life values. The larger overshoot and slower stabilization observed for longer half-lives suggest that weak degradation conditions introduce greater parameter sensitivity and increase the difficulty of inverse identification.

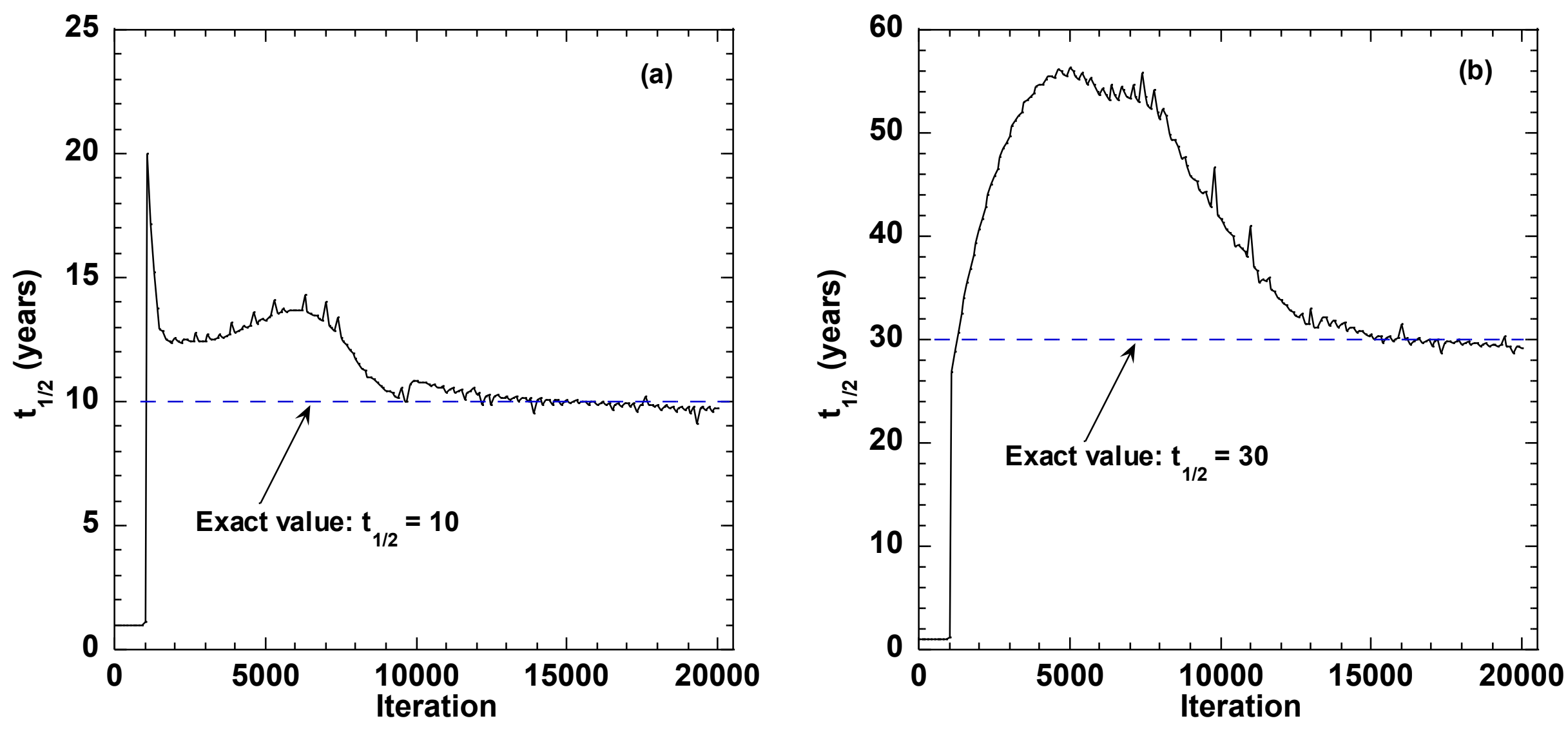

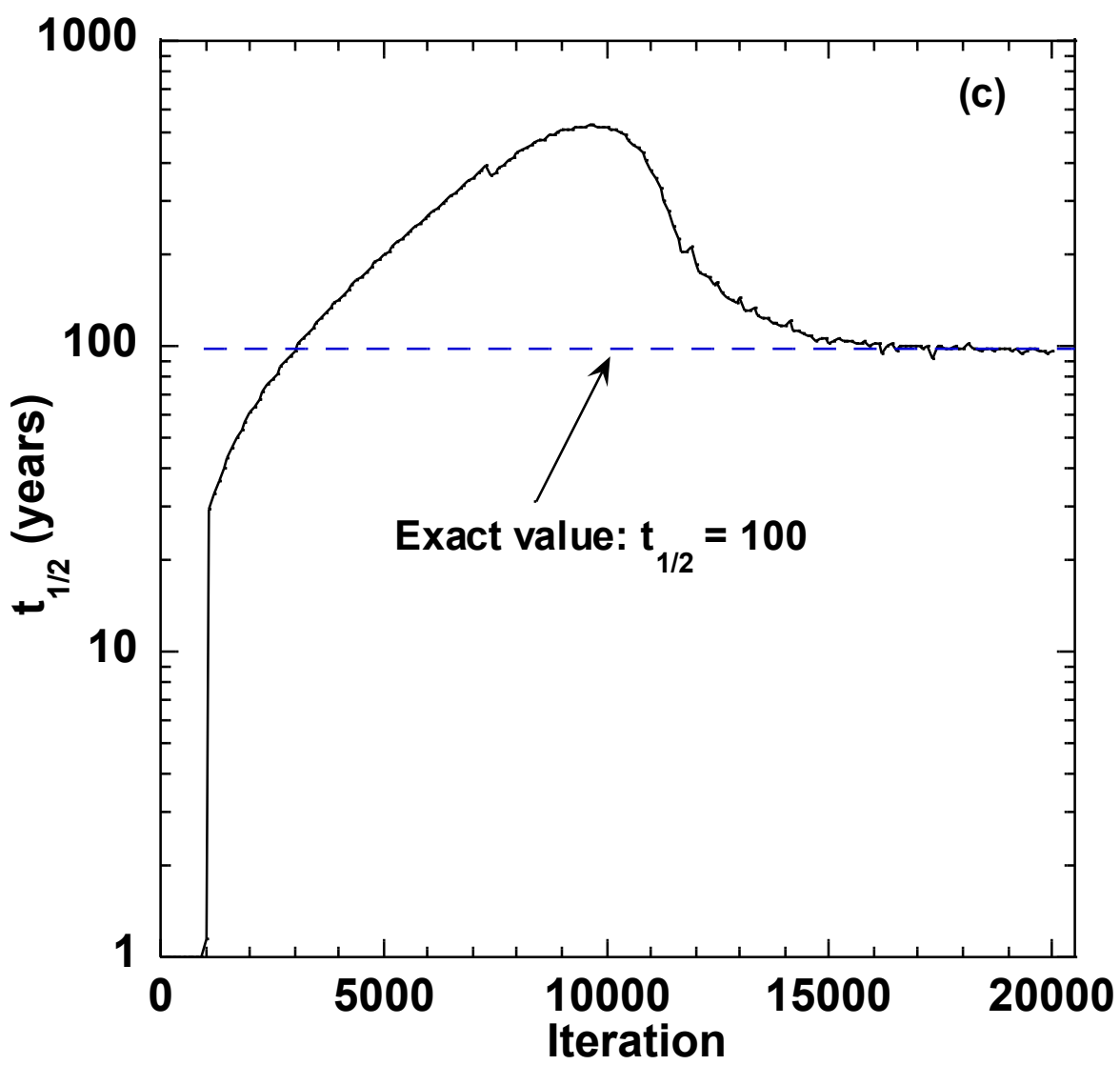

**Fig. 11.** Convergence histories of the inversely identified soil liner degradation half-life under $H_w = 3.0$ m: (a) exact $t_{1/2}$ =10 years, (b) exact $t_{1/2} = 30$ years, and (c) exact $t_{1/2} = 100$ years.

The robustness of the inverse PINN framework was further evaluated by introducing different levels of observation noise into the training data. Table 3 summarizes the inverted SL half-life values and their corresponding relative errors for exact half-life of 10, 30, and 100 years. As shown in Table 3 and Fig. 12, the relative error of the identified soil liner half-life remains relatively small when the noise magnitude is low, particularly for noise levels up to approximately 20%. For all three target half-life cases, the relative errors remain below approximately 6.1% under noise-free to 20% noise conditions, indicating that the inverse model can tolerate a certain degree of data uncertainty. However, when the noise magnitude increases to 30% and 50%, the relative error increases noticeably. At 30% noise, the relative errors increase to 10.3%, 11.7%, and 17.2% for exact half-life of 10, 30, and 100 years, respectively. At 50% noise, the errors further increased to 31.3%, 29.2%, and 27.3%, respectively. These results suggest that strong measurement noise can reduce the reliability of parameter identification. This trend is particularly evident at higher noise levels, where the inferred half-life becomes more sensitive to perturbations in the observed concentration data. Overall, the results indicate that the proposed inverse PINN framework exhibits acceptable robustness under low-to-moderate noise levels, while excessive noise may significantly affect the accuracy of the estimated degradation parameter.

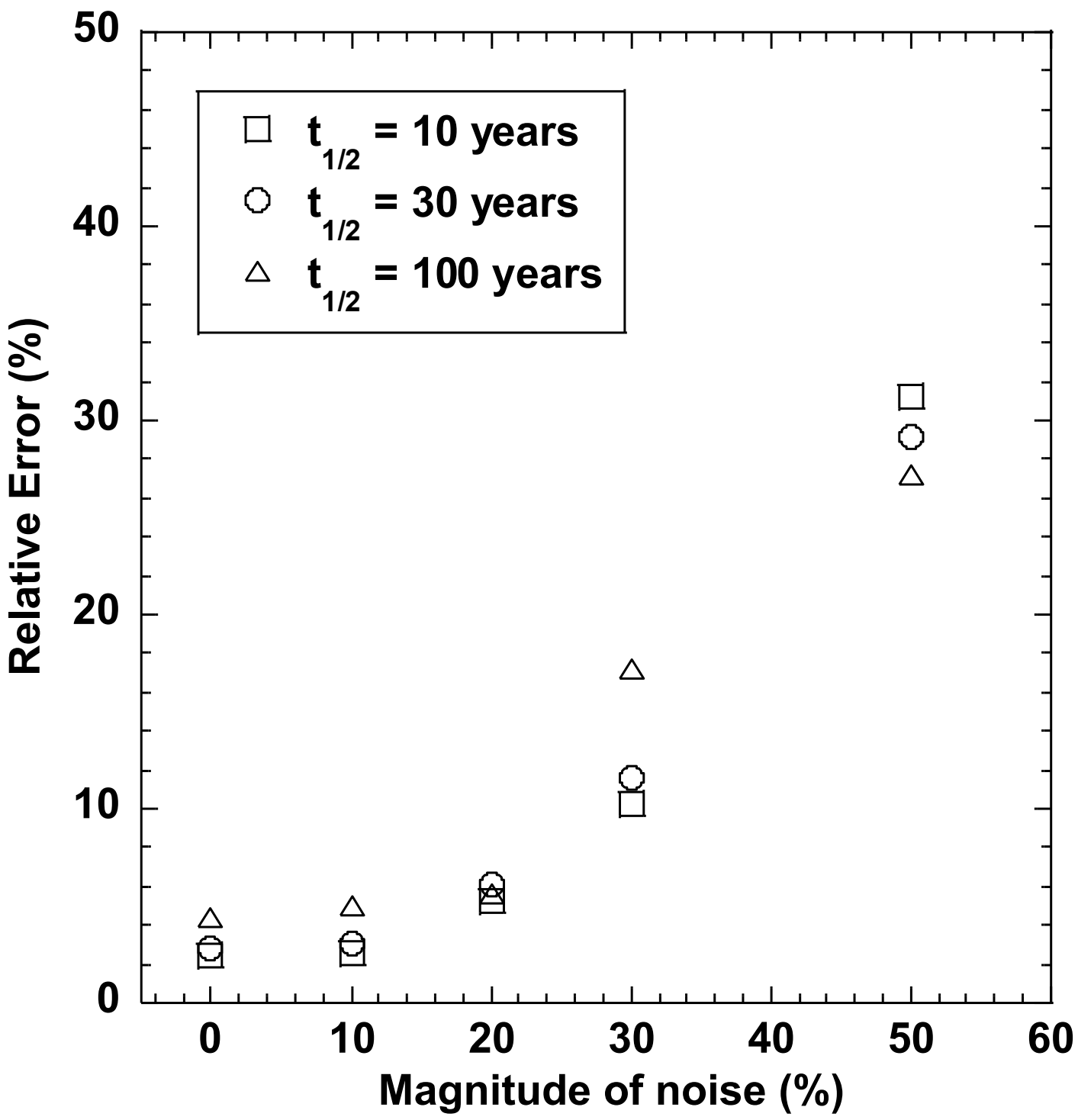

**Fig. 12.** Relative error of the inversely identified soil liner degradation half-life under different magnitudes of observation noise when $H_w = 3.0$ m.

**Table 3** Inverted values and their corresponding relative errors ($H_w = 3.0$ m)

| Noise Level (%) | Exact $t_{1/2}$ =10 years | | Exact $t_{1/2}$ =30 years | | Exact $t_{1/2}$ =100 years | |
|---|---|---|---|---|---|---|
| | Estimated $t_{1/2}$ SL (years) | Relative error (%) | Estimated $t_{1/2}$ SL (years) | Relative error (%) | Estimated $t_{1/2}$ SL (years) | Relative error (%) |
| Noise-free (0) | 9.8 | 2.5 | 29.2 | 2.8 | 104.4 | 4.4 |
| 10 | 9.7 | 2.6 | 30.9 | 3.1 | 95.0 | 4.9 |
| 20 | 10.5 | 5.3 | 31.8 | 6.1 | 105.6 | 5.6 |
| 30 | 11.0 | 10.3 | 33.5 | 11.7 | 82.8 | 17.2 |
| 50 | 6.88 | 31.3 | 38.8 | 29.2 | 127.3 | 27.3 |

# 6. Summary and Conclusion

This study developed a physics-informed neural network framework for modeling biodegradable contaminant transport through a two-layer GCL/SL composite liner system. The governing advection-dispersion-biodegradation equations, boundary conditions, and GCL/SL interface-continuity conditions were incorporated into the PINN formulation. Both a standard soft-constrained PINN (Std-PINN) and a hard-constrained PINN (H-PINN) were evaluated for forward prediction, and the hard-constrained PINN was further extended to inverse identification of degradation half-life. Based on the results obtained in this study, the following conclusions can be drawn:

1. The proposed two-network PINN framework can effectively represent contaminant transport through the GCL and SL layers, where the two materials have distinct thicknesses and transport properties. By using separate neural networks for the GCL and SL, the model can explicitly account for the layered structure of the composite liner and enforce concentration and flux continuity at the material interface.

2. The H-PINN provides more stable and accurate predictions than the Std-PINN. By embedding the prescribed concentration boundary condition and initial condition directly into the trial solutions, the hard-constrained formulation reduces the optimization burden associated with boundary- and initial-condition penalties. As a result, the model can focus more effectively on satisfying the governing-equation residuals and interface-continuity conditions.

3. The selection of activation function and neural-network structure has a clear influence on model performance. Among the activation functions considered, the Tanh function provides the best agreement with the reference solution and was therefore selected for subsequent simulations. The network-structure study indicates that increasing model capacity improves prediction accuracy up to an optimal range, while excessively deep or wide networks provide limited additional benefit and may increase optimization difficulty.

4. The inverse-modeling results demonstrate that the proposed H-PINN framework can successfully identify the degradation of half-life of the soil liner from limited concentration observations. The inferred half-life values converge toward the prescribed exact values under different target half-life cases. The noise-sensitivity analysis shows that the inverse PINN framework has acceptable robustness under low-to-moderate observation noise. The relative error of the identified half-life remains relatively small when the noise level is moderate but increases noticeably when the noise magnitude reaches high levels.

Although the proposed H-PINN framework demonstrates improved accuracy and stability compared with the Std-PINN model, several limitations should be acknowledged. The present study considers an idealized one-dimensional GCL/SL composite liner system with homogeneous or piecewise-homogeneous material properties. In actual landfill liner systems, spatial variability in hydraulic conductivity, diffusion coefficient, porosity, retardation factor, and degradation behavior may influence contaminant migration. In addition, the current formulation assumes prescribed material parameters and simplified transport mechanisms, while coupled hydro-chemo-mechanical effects, chemical compatibility of the GCL, clogging, desiccation, and long-term aging of liner materials are not explicitly considered.

Future studies should extend the proposed framework in several directions. Experimental column-test data or field monitoring data should be incorporated to further evaluate the practical applicability of the H-PINN model for real composite liner systems. The governing equations can also be expanded to account for multi-species reactive transport, nonlinear adsorption, time-dependent degradation, and concentration-dependent hydraulic and diffusion properties. In addition, future work may consider two-dimensional or three-dimensional contaminant migration, including defects, wrinkles, overlap zones, and nonuniform leachate heads, which are commonly encountered in practical liner applications.

# Credit Author Statement

Dong Li: Methodology, Formal analysis, Investigation, Writing the original draft, Reviewing and editing the manuscript; Yapeng Cao: Reviewing and editing the manuscript, Funding acquisition, Supervision, Methodology; Haiping Fu: Conceptualization, Reviewing and editing the manuscript, Validation; Shutong Han: Visualization, Reviewing and editing the manuscript.

### Competing Interests Statement

The authors declare there are no competing interests.

# Data Availability Statement

Data generated or analyzed during this study are available from the corresponding author upon reasonable request.

# Acknowledgements

The authors gratefully acknowledge the support of National Natural Science Foundation of China (Grant No. 42401176).